\documentclass[sigconf]{acmart}

\usepackage[normalem]{ulem}
\useunder{\uline}{\ul}{}
\usepackage{tabularx}
\usepackage{enumitem}
\usepackage{xcolor}
\usepackage{colortbl}
\usepackage{multirow}
\usepackage{graphicx}
\usepackage{array}
\usepackage{CJKutf8}
\usepackage[export]{adjustbox}
\usepackage{subfigure}
\AtBeginDocument{%
	}
 
\setcopyright{acmcopyright}
\copyrightyear{2018}
\acmYear{2018}
\acmDOI{XXXXXXX.XXXXXXX}

\acmConference[Conference acronym 'XX]{Make sure to enter the correct
	conference title from your rights confirmation emai}{June 03--05,
	2018}{Woodstock, NY}
\acmPrice{15.00}
\acmISBN{978-1-4503-XXXX-X/18/06}

\newcommand{\checker}[0]{\includegraphics[width=0.025\textwidth]{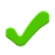}} 
\newcommand{\partchecker}[0]{\includegraphics[width=0.025\textwidth]{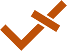}} 
\newcommand{\notchecker}[0]{\includegraphics[width=0.025\textwidth]{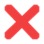}}

\begin{document}
	
	\title{ChiQA: A Large Scale Image-based  Real-World Question Answering Dataset for Multi-Modal Understanding}
	\author{Bingning Wang, Feiyang Lv, Ting Yao, Yiming Yuan}
	\author{Jin Ma, Yu Luo and Haijin Liang}
	\affiliation{%
		\institution{Tencent Inc.}
		\country{Beijing, China}
	}
	\email{bryantwwang,feiyanglv@tencent.com}
	
	
	\begin{abstract}
		Visual question answering is an important task in both natural language and vision understanding. However, in most of the public visual question answering datasets such as VQA \cite{antol2015vqa} CLEVR \cite{johnson2017clevr}, the questions are human generated that specific to the given image, such as `What color are her eyes?'. The human generated crowdsourcing questions are relatively simple and sometimes have the bias toward certain entities or attributes \cite{agrawal2018don,sugawara2018makes}.
		
		In this paper, we introduce a new question answering dataset based on image-ChiQA. It contains the real-world queries issued by internet users, combined with several related open-domain images. The system should determine whether the image could answer the question or not. Different from previous VQA datasets, the questions are real-world image-independent queries that are more various and unbiased. Compared with previous image-retrieval or image-caption datasets, the ChiQA not only measures the relatedness but also measures the answerability, which demands more fine-grained vision and language reasoning.
		
		ChiQA contains more than 40K questions and more than 200K question-images pairs.  A three-level 2/1/0 label is assigned to each pair indicating \textit{perfect answer}, \textit{partially answer} and \textit{irrelevant}. Data analysis shows ChiQA requires a deep understanding of both language and vision, including grounding, comparisons, and reading. We evaluate several state-of-the-art visual-language models such as ALBEF \cite{li2021align}, demonstrating that there is still a large room for improvements on ChiQA.
		
	\end{abstract}

	
	\maketitle
	
	\section{Introduction}
	Question Answering (QA) is one of the long-standing problems in natural language undertanding \cite{phillipsartificial,greenwood2005open}. Recently, with the introduction of large-scale QA datasets such as SQuAD \cite{rajpurkar2016squad} and Natural Questions \cite{kwiatkowski2019natural}, combined with the development of large-scale pre-training language models such as BERT \cite{devlin2019bert} and GPT-3 \cite{brown2020language}, etc, some of the state-of-the-art machine reading comprehension models even surpass human performance\footnote{\url{https://rajpurkar.github.io/SQuAD-explorer/}}. 
	
	\begin{figure}
		\centering
		\includegraphics[width=1.0\linewidth]{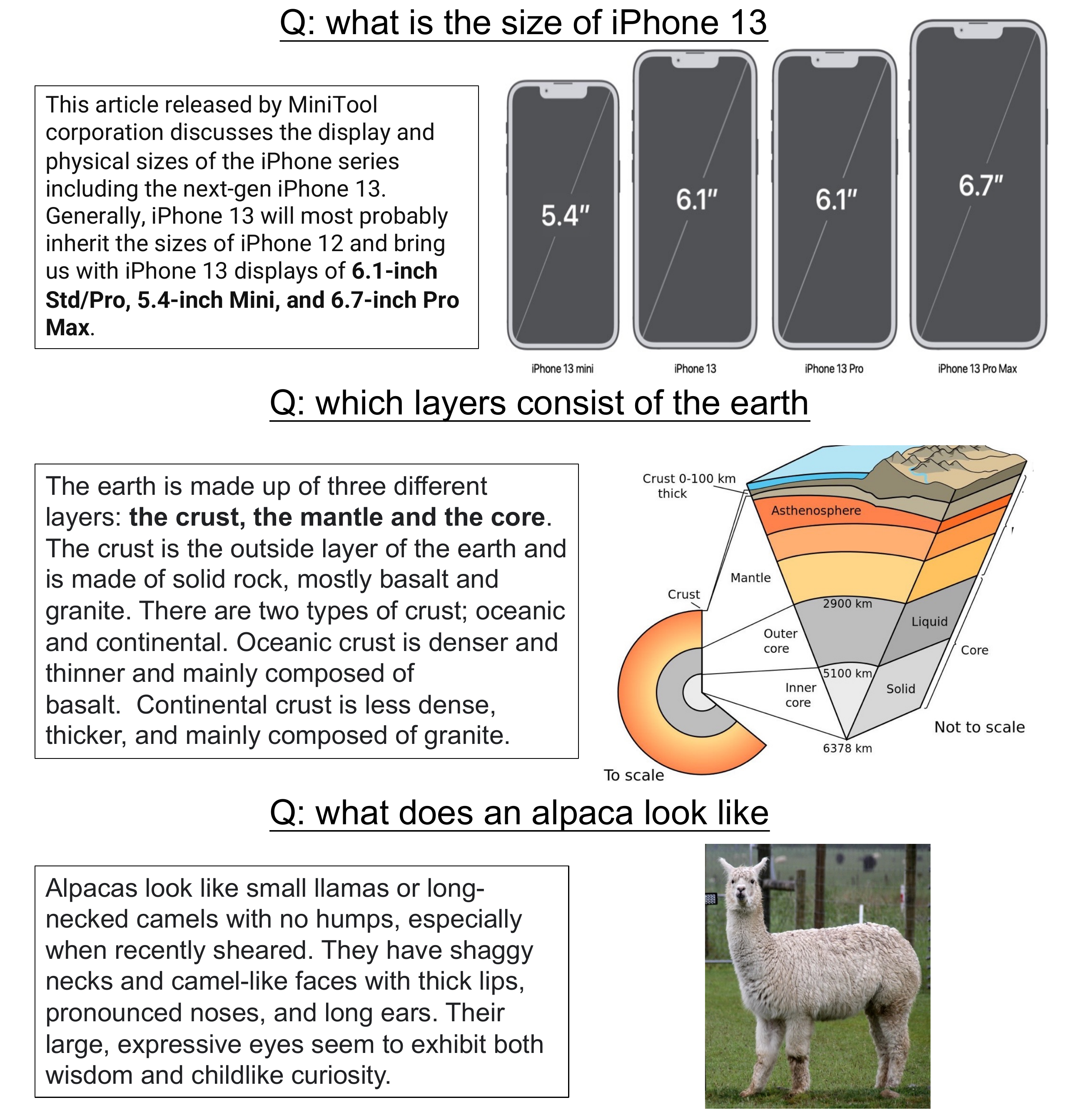}
		\caption{Examples show that for some questions, image answer is more clear and satisfactory than text-based answers. The left part is the text answer while the right side is the image answer. }
	\end{figure}
	
	Most of the QA datasets are mono-modal, where the questions, the answers and the resources (unstructured text or structure knowledge base) are text-based \cite{rajpurkar2016squad,kwiatkowski2019natural,berant2013semantic}.  However,  from the perspective of cognitive science \cite{maccabe1976theory,kumar2013system} and real-world applications \cite{hu2021question}, the image sometimes satisfies the user's demand more straightforwardly and conveniently.  For example, to answer the question ``\textit{what is the size of iPhone 13}'', an image illustrating the shape of four different versions of iPhone13 is more clear and more intuitive than a detailed text description. Therefore, it becomes more and more important to incorporate the image in question answering.
	
	A lot of models and datasets have been proposed to tackle the multi-modal visual question answering problem, such as VQA-1.0 and v2.0 \cite{antol2015vqa,goyal2017making}, CLEVR \cite{johnson2017clevr}, and GQA \cite{hudsongqa}, etc. In most of those datasets, a synthetic or real-world image is provided and human annotators are asked to generate questions with specific focuses, such as entities, relations, or both. However, there are several limitations of these previous VQA datasets. First of all, most of the previous VQA datasets are image-dependent, that is, the questions are asked \textit{after} they see the image, such as a question starts with ``Do you see a ... ''. A shortcoming of a context-dependent dataset is that the questions sometimes lack diversity, and models could easily learn the linguistic features merely based on the question and give the answer without understanding the context \cite{sugawara2018makes,goyal2017making}. Efforts to address this problem have mainly focused on making the datasets more balanced and diverse \cite{goyal2017making}. Nevertheless, benchmarks that involve crowdsourced question generation are bound to encode certain cognitive and social biases \cite{agrawal2018don,shrestha2022investigation}. In addition, the human-generated questions may merely be used for evaluating the model, but they are hard applied to a real-world problem.
	
	Secondly, most of the answers in previous VQA datasets are text-based entities, relations, or explanations, which are associated with local regions in the images. However,  for a question like ``what does an alpaca look like'', a single image illustrating an alpaca could perfectly serve as the answer. Further providing a complicated description of the alpaca's appearance is difficult and unnecessary. In addition, short-form text answers sometimes restrict the annotator to focus on local structure but not the global information \cite{fan2019eli5,kwiatkowski2019natural,wang2021comqa}. Therefore, a text answer may not be indispensable in the scenario of information seeking VQA.
	
	Furthermore, most of the previous VQA datasets are developed in English. Efforts on multi-modal learning in other languages are mainly focused on cross-lingual representation \cite{gao2015you,huang2021multilingual}. We argue that sometimes the linguistic information for one language is hard to transfer to other languages in VQA. For example, the \textit{pineapple} has two references in Chinese: \begin{CJK}{UTF8}{gbsn}`菠萝'\end{CJK} and \begin{CJK}{UTF8}{gbsn}`凤梨'\end{CJK}\footnote{https://zh.wikipedia.org/zh-hans/\%E8\%8F\%A0\%E8\%98\%BF, this two species of pineapple only grown in Southeast Asia, sometimes they are deemed as the same fruit but actually they have a minor difference.}, when we asked in Chinese about their difference  \begin{CJK}{UTF8}{gbsn}`菠萝和凤梨的区别是什么'\end{CJK}, little cross-lingual information could be borrowed from other languages. Therefore, developing a specific VQA dataset for a specific language is also important.
	
	In this paper, we propose a new \textbf{Ch}inese \textbf{i}mage \textbf{q}uestion \textbf{a}nswering dataset called ChiQA. In ChiQA, given a real-world query and several related images, a three-level 2/1/0 label is assigned to each image indicating whether they are \textit{perfect answer}, \textit{partially answer} or \textit{irrelevant}. There are three main characteristics of ChiQA:
	
	\begin{itemize}[wide, labelwidth=!, labelindent=0pt]
		\setlength{\itemsep}{0cm}
		\setlength{\parsep}{0.0cm}
		\setlength{\parskip}{0.0cm}  
		\item  \textbf{Real world question}: The questions in ChiQA are open-domain user queries issued to search engine. Real-world queries are diverse since they convey the user's demand for specific information and thus would be either factoid or non-factoid, spanning many types of domains such as medical, health, etc. In addition, real-world queries are image-independent which would be more balanced and unbiased.
		\item   \textbf{Answerability}: The images in ChiQA are also real-world images collected from search engine, which are related to but not necessarily answerable for the question. Answerability is very important and difficult for question answering, it demands a deep understanding of both the question and the context \cite{yang2015wikiqa,jia2017adversarial,nakanishi2018answerable}.
		\item  \textbf{Unbiased}: Our data crowdsourcing involves a two-phase active learning process. In the first phase, we randomly collect the samples from the web. In the second phase, we first train the model based on the data from the first phase, then we use the trained model to choose the \textit{hard} examples on the remaining data and continue labeling. This two-phase setting makes the data more challenging \cite{swayamdipta2020dataset}, and inherently debiasing the unreasonable favoritism towards certain attributes or linguistic patterns in the data \cite{zhuang2015debiasing,paullada2021data}. 
	\end{itemize}
	
	\begin{figure}
		\centering
		\includegraphics[width=1.0\linewidth]{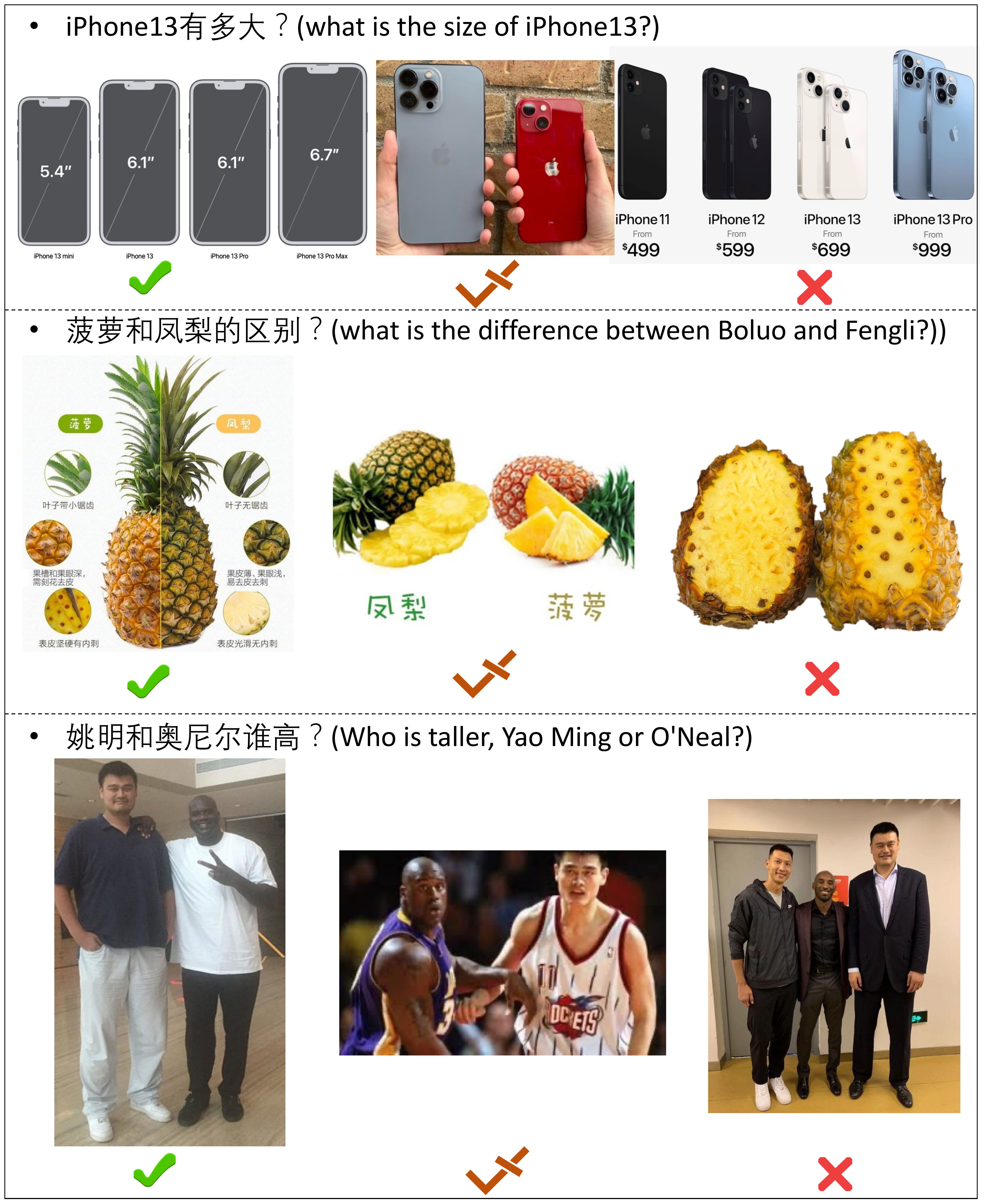}
		\caption{Examples of the proposed ChiQA dataset. For each question, several related images are labelled with 2-1-0 scores indicating whether they are answer (\checker), not the answer(\notchecker) or the partially answer (\partchecker) to the question.}\label{chiqa.sample}
	\end{figure}

	The collection on ChiQA consists of a rigorous inspection step to maintain the quality. We collected more than 40,000 queries with more than 200,000 question-image pairs. Some examples of ChiQA are shown in Figure \ref{chiqa.sample}. Qualitative analysis of the data shows that ChiQA requires a deep understanding of both language and vision, including grounding, reading, logical inference, etc.
	We evaluate several state-of-the-art visual-language models such as ALBEF \cite{li2021align}, which are far behind human performance, demonstrating that there is still large room for improvements on ChiQA.
	
	\renewcommand\arraystretch{1.1} 
	\begin{table*}[!ht]
		\begin{tabular}{ccccccc}
			Dataset              & Question Source      & Image Source         & Answer Type          & Size                 & Domain               & Language             \\ \noalign{\hrule height 1.15pt}
			DAQUAR               & Crowdsourcing        & NYU-DEPTH-V2         & Objects/Color/Number & 12,468               & Close        & English              \\ \hline
			CLEVR                & Synthesis            & Synthesis            & Objects/YesNo/Number & 999,968              & Close            & English              \\ \hline
			VQA-1.0                & Crowdsourcing            & MS-COCO \& Abstract Scenes            & Objects/YesNo/Number & 614,163              & Open            & English              \\ \hline
			VQA-2.0                & Crowdsourcing            & MS-COCO           & Objects/YesNo/Number & 1,105,904              & Open            & English              \\ \hline
			Visual Genome                & Crowdsourcing            & MS-COCO           & Open-ended & 1,445,322             & Open            & English              \\ \hline
			Visual7W                & Crowdsourcing            & MS-COCO           & Multiple-choice &  327,939             & Open            & English              \\ \hline
			Visual Madlibs                & Crowdsourcing            & MS-COCO           & Cloze &  360,000             & Open            & English              \\ \hline
			KB-VQA                & Crowdsourcing            & MS-COCO           & Objects/YesNo/Number  &  2,402             & Close            & English              \\ \hline
			FVQA                & Crowdsourcing            & MS-COCO \& ImageNet          & Entity &  5,826             & Open            & English              \\ \hline
			FM-IQA               & Crowdsourcing            & MS-COCO           & Open-ended &  316,193            & Open            & Chinese              \\ \hline
			TextVQA & Crowdsourcing & Open Images & Open-ended &45,336 & Open & English \\ \hline
			MovieQA               & Crowdsourcing            & Plot Synopses           & Multiple-choice &  14,944            & Movie            & English  \\ \hline
			\rowcolor[HTML]{D9D9D9}
			ChiQA               & User Queries            & Web          & Score &       210,681       & Open           & Chinese              \\ \noalign{\hrule height 1.15pt}		
		\end{tabular}
		\caption{A summary of major VQA datasets and their main characteristics. The question in our proposed ChiQA is a real world query that is not specific to the target image. And the images are obtained from the open-domain web.}\label{table.dataset}
	\end{table*}
	
	\section{Related Work}
	\subsection{Question Answering}
	Question Answering has been a long-term focus in NLP that can date back to 1960s, when \citet{green1961baseball} proposed a system to answer questions about baseball games. Since then, many works have been done to use diverse data resources to answer any type of question. Recently, due to the development of large-scale dataset SQuAD \cite{rajpurkar2016squad}, and the pretraining language models such as BERT \cite{devlin2019bert} and GPT-3 \cite{brown2020language} in NLP, the performance on QA has been greatly improved and even surpass the human. However, since the questions in SQuAD are human-created which contains some unfavorable bias \cite{sugawara2018makes}, recent efforts in QA has focused on open-domain real-world questions, such as Natural Questions \cite{kwiatkowski2019natural}, MS-MARCO \cite{nguyen2016ms}, Dureader \cite{he2018dureader} etc. Among them, our proposed ChiQA is very related to WikiQA \cite{yang2015wikiqa}, it maintains more than 3,000 questions, of which about two-thirds are not answerable questions since they never have a positive sentence in the corresponding paragraphs. However, our proposed ChiQA is multi-media and large-scale.
	
	\subsection{Visual Question Answering}
	There have been years of effort in connecting the visual and textual information for joint multimodal learning \cite{barnard2003matching,zitnick2013learning,karpathy2015deep}. \citet{geman2015visual} introduced a restricted visual Turing test to evaluate visual and language understanding. \citet{malinowski2014multi} proposed the DAQUAR, which is a small size VQA dataset built upon indoor scene RGB-D images. Currently, most of the VQA datasets are built based on MS-COCO dataset \cite{lin2014microsoft} which provides both images and captions. \citet{antol2015vqa} introduce the task of free-form and open-ended VQA which takes the image and a free-form, open-ended, natural-language question about the image as input, and produces a
	natural-language answer as the output.  VQA2.0 \cite{goyal2017making} and CLEVR \cite{johnson2017clevr} datasets are designed to address bias and reasoning limitations of VQA1.0, respectively. Visual Genome \cite{krishna2017visual} is a very large-scale VQA dataset built with images from the Visual Genome project, it includes structured annotations of scene graphs describing the visual elements of the scenes, their attributes, and relationships between them. Visual7w \cite{zhu2016visual7w} is a subset of the Visual Genome, specifically, all the objects mentioned in the questions are visually grounded with bounding boxes of their depictions in the images. \citet{wang2017explicit} proposed KB-VQA which requires employing external knowledge rather than learning a mapping from image and question. \citet{tapaswi2016movieqa} introduce the MovieQA to answer the question about a movie, it requires multiple sources of information such as video clips, plots, subtitles, etc. Recently, \citet{singh2019towards} proposed text-VQA, where the task is to answer human-generated questions about the embedded text in the image.
	
	Based on these datasets, a lot of cross-modality frameworks have been proposed to solve this problem, such as modular neural networks \cite{andreas2016neural,yi2018neural}, attention-based networks \cite{tan-bansal-2018-object,hudson2018compositional}. Recently, inspired by the language model pretraining in NLP, some cross-modality vision and language pre-training methods have also been proposed \cite{su2019vl,tan2019lxmert,li2020oscar,li2021align,gu2022wukong}. Those pre-training methods utilize large-scale, probably noisy, image and caption pairs and achieve state-of-the-art results on various VQA problems.
	
	\subsection{Image Retrieval}
	Image retrieval is also related to our proposed ChiQA. In this task, we are given a text, and the system should retrieve related images with respect to the input text from the images pool, or verse visa. Two widely used image retrieval datasets are Flickr30K \cite{flickr30k} and MS-COCO \cite{lin2014microsoft}. The efforts to improve the performance is focused on large-scale contrastive representation learning \cite{qi2020imagebert,li2021align,radford2021learning}. However, in image retrieval, the text is sometimes the descriptive caption of the corresponding image that can be viewed as a sort of \textit{translation}. While in VQA the text and image are non-parallel that the text may only ask about a small part of the image. Therefore, ChiQA is a more fine-grained task compared with image retrieval.
	
	\subsection{Image Captioning and Visual Entailment}
	Image caption is a popular research area of visual and language understanding that deals with image understanding and a language description for that image. MS-COCO \cite{lin2014microsoft} is the most popular benchmark dataset of image captioning.  Inspired by the sequence-to-sequence model for machine translation, most captioning models mainly adopt an encoder-decoder
	framework, where an encoder network encodes the input image into an embedding space and a decoder network takes the encoder feature as input and generates the output caption \cite{vinyals2016show,rennie2017self,anderson2018bottom}. Some recent works also utilize large-scale pretraining which greatly improves the performance on image captioning \cite{zhou2020unified,li2020oscar,hu2021unit,huo2021wenlan}.
	
	Visual Entailment is the task to predict whether there exists some relationship between the images and texts, which is inspired by the natural language inference task \cite{bowman2015large}. \citet{xie2019visual} design an inference task where the \textit{premise} is image and the system should predict whether it could entail/contradict the \textit{hypothesis} text.  \citet{suhr2019corpus} proposed the NLVR2 in which the premise consists of two images and the task is to determine whether a caption is true regard to the premise image pairs. 
	
	Our proposed ChiQA focuses on real-world multi-modal learning in which the questions are user queries issued to search engine.  In addition, we did not utilize MS-COCO as our image source, since it is object-based \cite{lin2014microsoft} where the images are obtained by a closed set of object categories such as `dog', `banana', etc. It is not as diverse as the open domain web images.
	We summarize the key characteristics of the different VQA datasets in Table \ref{table.dataset}.

	\begin{figure}
		\centering
		\includegraphics[width=0.95\linewidth]{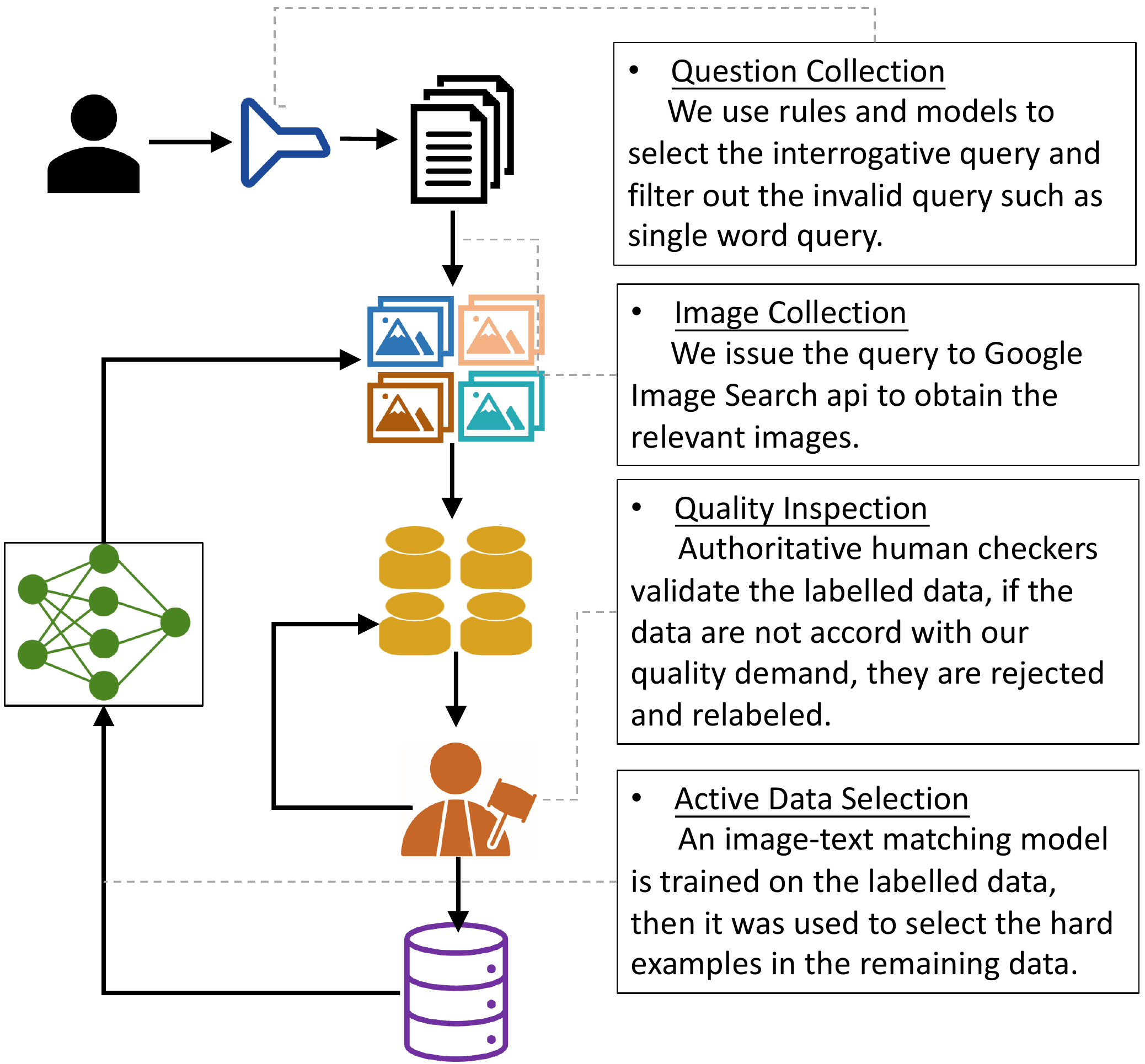}
		\caption{The labeling process of the ChiQA dataset. First of all, we use rules and models to select the question from the user queries issued to search engine. Then we adopt Google Image search API to obtain the image URL and then crawl the corresponding image. Then the question, images are labeled by the crowdsourcing workers to obtain labels. Next, the labeled samples are checked by an authoritative checker to maintain the quality of the data. Finally, an active learning  process is applied to the labeled data and used to select the remaining data to be labeled.}\label{label.process}
	\end{figure}
	
	\section{Data Collection}
	Each example in ChiQA contains a user query and five images. Each image is related to but not necessarily answerable to the question. We label each image with three-level labels, 2/1/0, to indicate their answerability. In addition, for each question-image pair, we also provided the text content of the web page where the images from.
	The whole data collection process is illustrated in Figure \ref{label.process}.
	
	\subsection{Question Collection}\label{sec.question.collection}
	We use the user queries issued to search engine as our question source. Since user's query contains a large amount of samples that have no question intent, such as searching for a specific entity, homepages, etc. We use a binary classification model trained in a weakly supervised way to determine whether a query is a question or not. To build such a model, we collect a large but weakly supervised training data in four ways: 1) we select the query that has a high click ratio of the community question answering (CQA) site, such as Quota, Zhihu, etc, as the positive query; 2) the title of the CQA dataset as the positive data; 3) inspired by Natural Questions \cite{kwiatkowski2019natural}, we use rules to select the positive query from the log data: a) whether the question contains the interrogative pronoun such as \begin{CJK}{UTF8}{gbsn}`谁是'\end{CJK}(`\textit{who}'), \begin{CJK}{UTF8}{gbsn}`如何'\end{CJK}(\textit{`how to'}), etc. b) the title containing the informative words such as \begin{CJK}{UTF8}{gbsn}`的过程'\end{CJK}(`\textit{the process of}'), \begin{CJK}{UTF8}{gbsn}`的方法'\end{CJK}(`\textit{the method to}'), etc; 4) the query that has high click ratio of the non-QA site, such as video site, homepage, novel site, etc, is treated as the negative sample.
	
	We use the weakly supervised data to train a simple bi-directional LSTM \cite{Hochreiter1997LongSM} model with binary cross-entropy loss. We found the model is simple yet the performance is satisfactory compared with a more heavyweight model like BERT. On the held-out human-labeled test set, our model achieves more than 90\% accuracy with more than 80\% recall. We use this model to select more than 75k queries to the next step.
	
	\subsection{Image Collection}
	After we obtain the question, we issued them to Google Image search api\footnote{https://serpapi.com/images-results}. The Google Image API returns 100 most related images for the input query and every image URL is validated by default. However, we found in some cases the image may not be available, so we use the in-house spider to crawl the images and dismiss those images that are unavailable. Furthermore, we only keep images with both dimensions greater than 200 pixels, and the ratio of large-to-small dimensions is no more than three. In this way, we filter out images that are too small or are very tall or wide, which can be of low-resolution after image augmentations like up-sampling.

	\begin{figure}
		\centering
		\includegraphics[width=0.92\linewidth,frame]{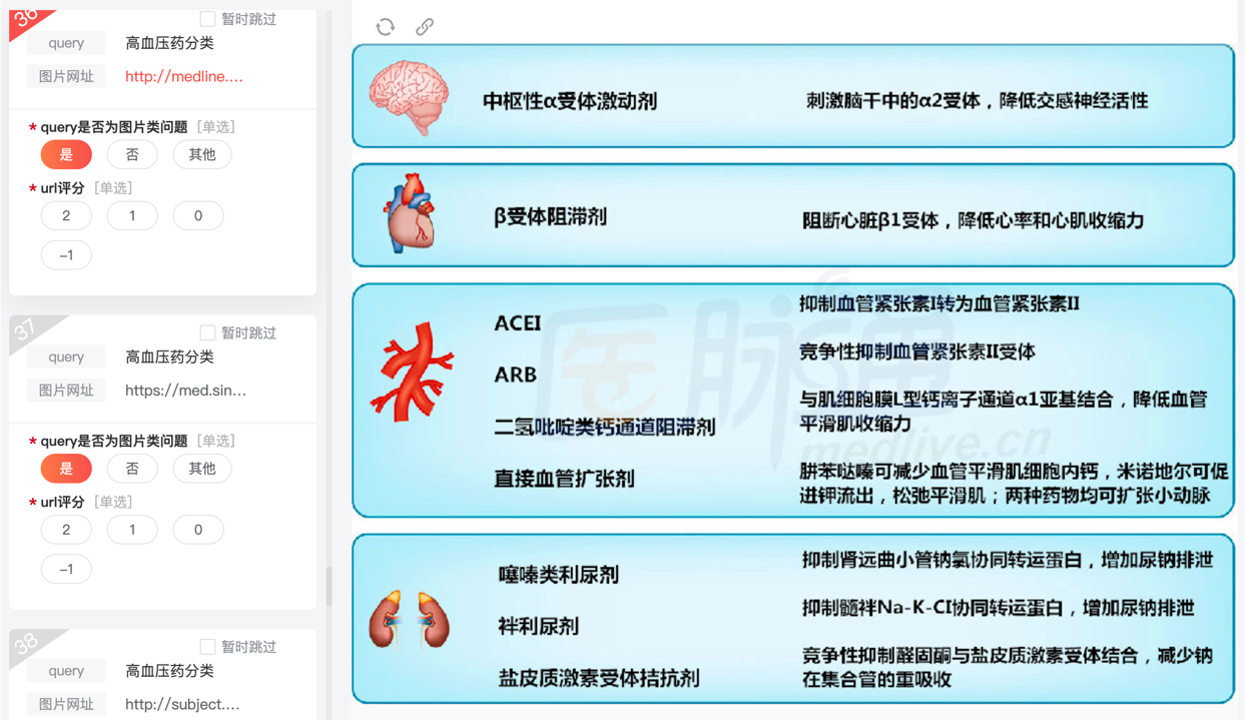}
		\caption{The annoation interface of ChiQA. The left panel is the label frame and the right is the image.}\label{label.interface}
	\end{figure}
	
	\subsection{Image Labelling}
	After we collected the question and corresponding five images, we ask crowdsourcing workers to label the answerability. Annotation is performed using a custom annotation interface in Figure \ref{label.interface}.  The guidelines and interface divide the annotation task into three conceptual stages, where all three stages are completed by a single annotator in a single session.
	
	\textbf{Question Identification:} Contributors of ChiQA determine whether the question could be answered by an image. A good image question is the one that itself could be answered by an image, such as  \begin{CJK}{UTF8}{gbsn}`xxx的差别'\end{CJK}(`\textit{what is the difference ...}'), the question that is ambiguous, incomprehensible or containing baseless statement are regarded as the invalid question and not involves in the subsequent process.  Note that 1) we regard the non-factoid opinion-seeking question as good since we believe for some questions like  \begin{CJK}{UTF8}{gbsn}`红色毛衣配什么裤子好'\end{CJK} (`What trousers does a red sweater go well with?') an image could satisfy the question well compared with the text answer. 2) The question that can be answered by a single entity, like \begin{CJK}{UTF8}{gbsn}`肖申克救赎的导演是谁'\end{CJK} (`Who is the director of The Shawshank Redemption?'), are regarded as \textit{bad}  question since it requires nothing but world knowledge and grounding. 3) For some queries that are falsely classified as a question by the model introduced in section \ref{sec.question.collection}, it should be given a label -1. 
	
	\textbf{Image Identification:}  For those questions that are labeled as \textit{good} (1), annotators then select whether the image is perfect, partially or unable to answer the question, with a three-level 2/1/0 label. For ChiQA, there is some intermediate zone where the answer image could be either right or wrong, so we add a level 1 to indicate the image could somewhat answer the question but the user who takes the image may conduct another query to confirm the answer. We did not adopt the quintile(1-5) labeling standard since we found in the preliminary process that this difference is too subtle to discriminate.
	
	Some guideline examples for labeling the question and image are shown in Table \ref{tabel.label.question}.

	\renewcommand\arraystretch{0.95} 
	
	\begin{table}[]
		\small
		\begin{tabularx}{0.43\textwidth}{|l|X|} 
			\multicolumn{2}{c}{	\begin{normalsize} Query Labelling\end{normalsize}} \\ \hline \hline
			\multicolumn{1}{|c|}{Query Type}        & \multicolumn{1}{c|}{Examples}                                  \\ \hline 
			& \begin{CJK}{UTF8}{gkai}太阳系八大行星的位置\end{CJK}                                       \\
			& 		\begin{footnotesize} \textcolor{gray}{what is the position of eight planets in the solar system.} \end{footnotesize}    \\ \cline{2-2} 
			& \begin{CJK}{UTF8}{gkai}中国的南北分界线是什么\end{CJK}                                                           \\
			& \begin{footnotesize} \textcolor{gray}{what is the dividing line between north and south china  } \end{footnotesize}      \\ \cline{2-2} 
			& \begin{CJK}{UTF8}{gkai}如何使用不同介词\end{CJK}                                                           \\
			& \begin{footnotesize} \textcolor{gray}{How to use prepositions    } \end{footnotesize}                \\ \cline{2-2} 
			& \begin{CJK}{UTF8}{gkai}圆脸女孩适合什么发型\end{CJK}                                                           \\
			\multirow{-8}{*}{valid question (1)}   & \begin{footnotesize} \textcolor{gray}{ What hairstyle makes a girl with round face cut looks better?  } \end{footnotesize}    \\ \hline
			& \begin{CJK}{UTF8}{gkai}城镇户口可以买农村旧宅房吗\end{CJK}                                                           \\
			& \begin{footnotesize} \textcolor{gray}{ Can urban households purchase old rural houses now?   } \end{footnotesize}         \\ \cline{2-2} 
			& \begin{CJK}{UTF8}{gkai}肖申克救赎的导演是谁\end{CJK}                                                           \\
			& \begin{footnotesize} \textcolor{gray}{ Who is the director of The Shawshank Redemption? } \end{footnotesize}       \\ \cline{2-2} 
			& \begin{CJK}{UTF8}{gkai}月亮上除了桂树还有什么植物\end{CJK}                                                           \\
			\multirow{-6}{*}{invalid question (0)} & \begin{footnotesize} \textcolor{gray}{What plants are there on the moon besides laurel trees? } \end{footnotesize}       \\ \hline
			& \begin{CJK}{UTF8}{gkai}今天天气真好\end{CJK}                                                           \\
			& \begin{footnotesize} \textcolor{gray}{ It's a beautiful day today.   } \end{footnotesize}           \\ \cline{2-2} 
			& \begin{CJK}{UTF8}{gkai}农历新年的礼物\end{CJK}                                                           \\
			\multirow{-4}{*}{not a question (-1)}  &\begin{footnotesize} \textcolor{gray}{ Chinese New Year gifts } \end{footnotesize}              \\ \hline		
			\multicolumn{2}{c}{\begin{normalsize} Image Labelling \end{normalsize}} \\ \hline
		\end{tabularx}
		\includegraphics[width=0.93\linewidth]{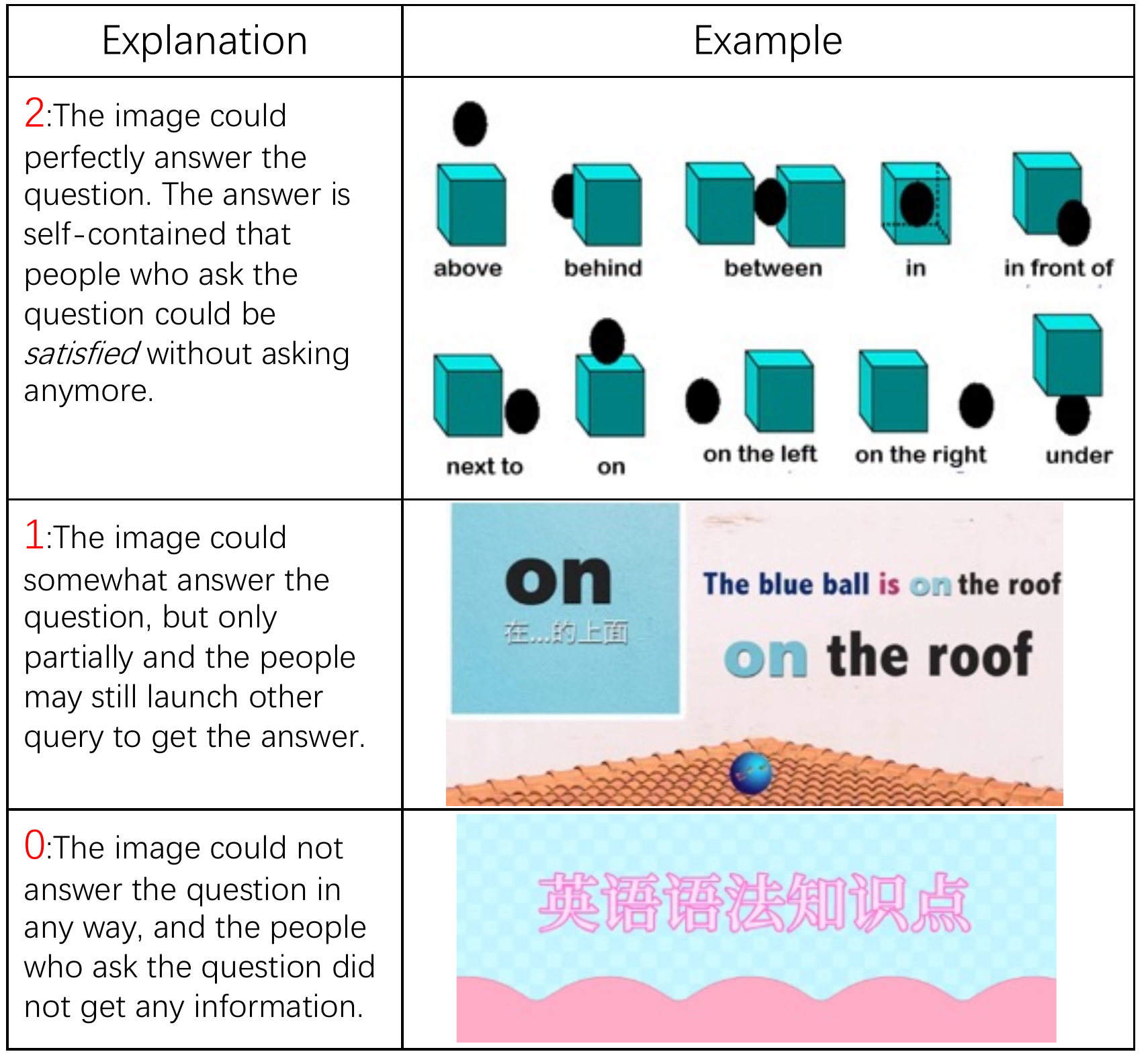}
		\caption{Some guideline examples to label the question and image. For the image, the question is \begin{CJK}{UTF8}{gbsn}如何使用不同介词\end{CJK}(How to use prepositions)}\label{tabel.label.question}
		\vspace{-0.3cm}
	\end{table}
	
	\subsection{Quality Inspection}
	We employ a very rigid data evaluation process to maintain the quality. First of all, all data are checked by authoritative \textit{experts} who are well-informed of the rules and could make a fair judgment of the data. For each batch of annotated data, the experts sample one-fifth of the data and annotate it again. If the data had less than 90\% accuracy, this batch of data is rejected and relabelled by the crowdsourcing worker. 
	
	Secondly, since we employ the outsourcing group as our workforce, the annotation standard and accuracy of different outsourcing groups may vary a lot. Therefore, instead of asking only one group of annotators, we ask three groups of annotators with the same expert checker. The three groups annotate independently and their annotated batched data are checked by the expert, after three rounds of annotation, we select the group that has better average accuracy judged by the expert, and the remaining data are annotated by this \textit{winner} group.
	
	\subsection{Active Data Selection}
	One lesson we learned from the previous data collection process is that the dataset involves crowdsourced samples is bound to encode certain cognitive and social biases \cite{goyal2017making,agrawal2018don,shrestha2022investigation}. To make our data challenging enough and unbiased, we adopt an active-learning based data collection process consisting of two phases. In the first phase, we sample data randomly and annotate them. Then, we use the labeled data obtained from the first phase to train an ALBEF \cite{li2021align} model, a state-of-the-art model on vision and language learning. This ALBEF model is utilized as our \textit{selector} to select the hard example in the next round: for a new sample, if the model's prediction has a very high entropy which means the selector has low confidence, so it is \textit{hard} for the current model and deserves to be labeled, otherwise, the sample may be so \textit{easy} since the model could make a confident prediction, making this sample not very valuable to be labeled \cite{swayamdipta2020dataset}.
	
	The active learning data selection process indeed makes the dataset more unbiased. Similar with some previous works \cite{antol2015vqa}, we found the labeled data in the first phase contains some undetectable fake patterns. For example, the question including the word \begin{CJK}{UTF8}{gbsn}`的技巧'\end{CJK} (the skill of) are labeled as the valid question but nearly all of the corresponding images are labeled as unanswerable. Consequently, the model trained on this dataset may disregard the image and predict a very low score only if the question contains that word. The active learning process makes this biased sample less likely to be annotated in the next round, making the data more balanced.

	\begin{figure}
		\centering
		\subfigure{
			\begin{minipage}[b]{0.4\linewidth}
				\includegraphics[width=1\linewidth]{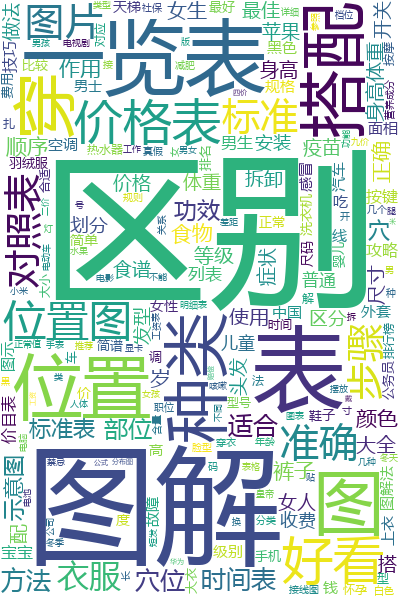}
			\end{minipage}
		}
		\subfigure{
			\begin{minipage}[b]{0.4\linewidth}
				\includegraphics[width=1\linewidth]{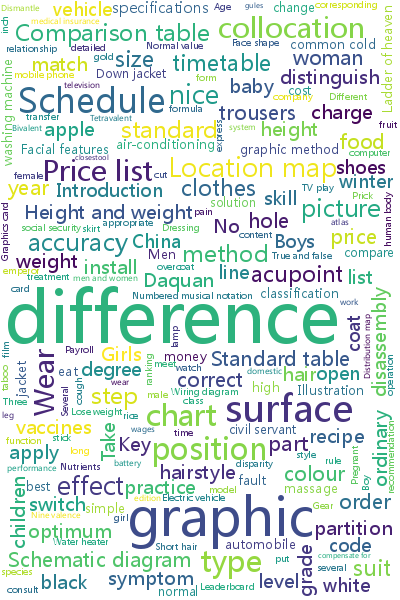}
			\end{minipage}
		}        
		\caption{Wordcloud of the questions in ChiQA. Words with larger size mean they appear more frequently. The left figure is the frequent words in ChiQA and the right figure is their English translation.}\label{fig.question.freq.words}
	\end{figure}

	\subsection{Test Set Collection}
	Since the test set demands a higher quality than the training set, we get the test set as follows: First of all, we randomly sample 2,500 samples from the training set and ask the crowdsource workers (probably different from the annotator that previously label this sample) to annotate them again. If the new labels are same as the old labels, they are passed to the test set; otherwise, we ask the authoritative experts to relabel them again. Finally, we obtain 2,362 test questions and more than 40,000 training questions, the statistics of ChiQA are shown in Table \ref{stat}:
	
		\begin{table}[!ht]
	\centering
	\small
	\setlength{\tabcolsep}{2.1pt}{
		\begin{tabular}{|c|c|c|c|c|c|}
			\hline
			& \#Questions & \#Images & \#2            & \#1             & \#0             \\ \hline
			Train & 40,635      & 198,440   & 35,771(18.03\%) & 73,660(37.12\%) & 89,009(44.85\%) \\ \hline
			Test  & 2,362       & 12,214    & 1,760(14.51\%)  & 3,604(29.41\%)  & 6,850(56.08\%)  \\ \hline
	\end{tabular}}
	\caption{The statistics of ChiQA. \#2 means the number of images that are labeled with score 2 in the dataset.}\label{stat}
\end{table}

	\section{Data Analysis}
	This section analysis our proposed ChiQA in three aspects: the question aspect, the image aspect, and the vision and language understanding aspect.
	
	\subsection{Question Analysis}
	\textbf{Frequent words:} we plot the most frequent words of our proposed ChiQA dataset in Figure \ref{fig.question.freq.words}, we can see that ChiQA contains various types of words such as  \begin{CJK}{UTF8}{gbsn}`区别'\end{CJK} (`the difference') and \begin{CJK}{UTF8}{gbsn}`位置'\end{CJK} (`position'), etc.
	
	\begin{figure}
		\includegraphics[width=1.05\linewidth]{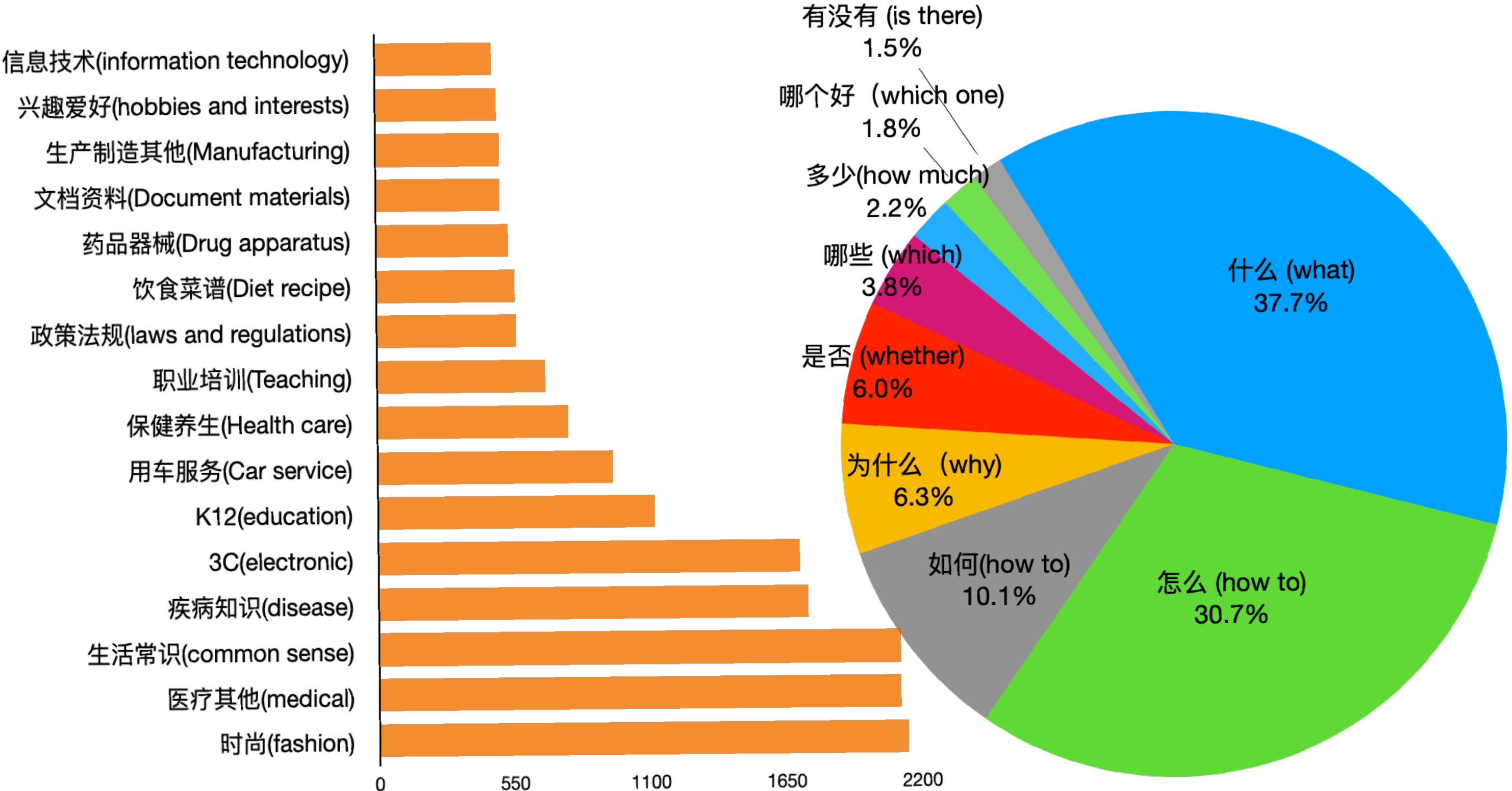}
		\caption{Left: The distribution of question domain, we only keep the top 16 most frequent classes. Right: The distribution of the interrogative words in ChiQA's question.}\label{label.question.class}
	\end{figure}

	\begin{figure*}
		\includegraphics[width=0.95\linewidth]{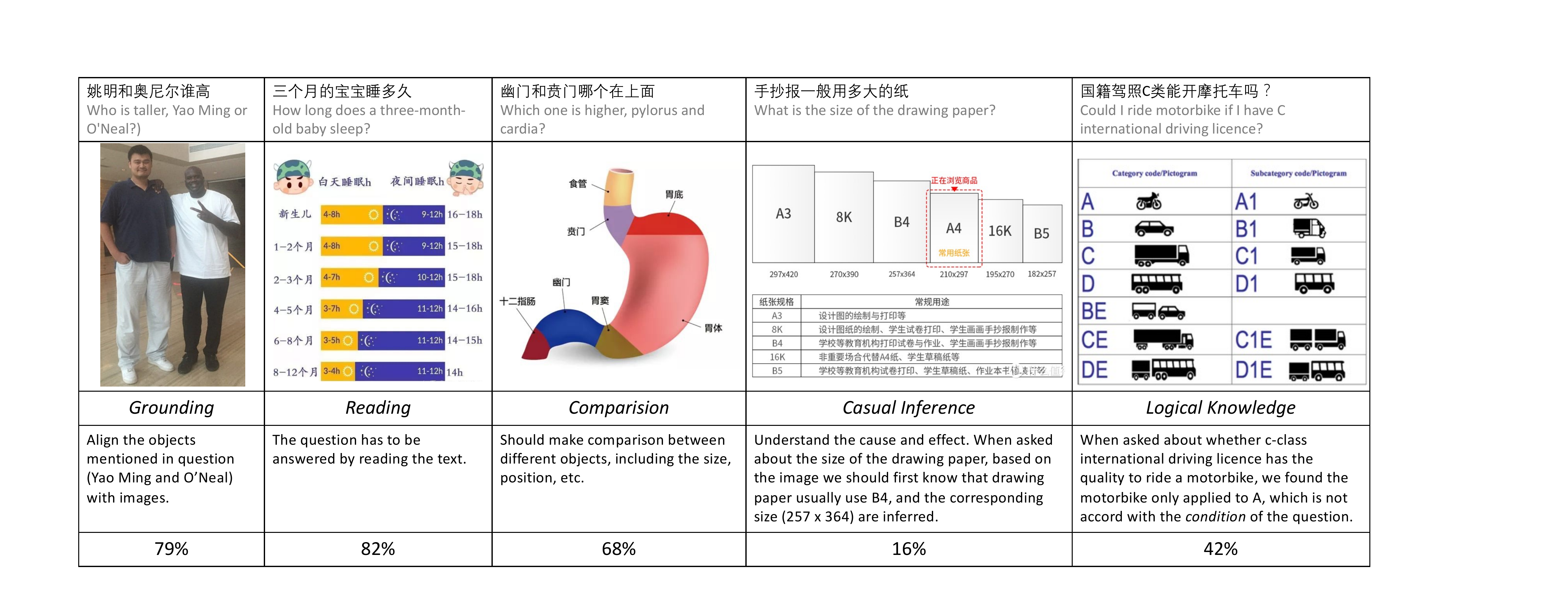}
		\caption{Prerequisite reasoning skills for ChiQA. From top to the bottom are the question, the image, the reasoning skill, the explanation,  and the proportion. Note that the reasoning skills are not mutually exclusive so some examples may require more than one reasoning skill.}\label{label.skill.class}
	\end{figure*}
	
	\begin{figure}
		\includegraphics[width=0.85\linewidth]{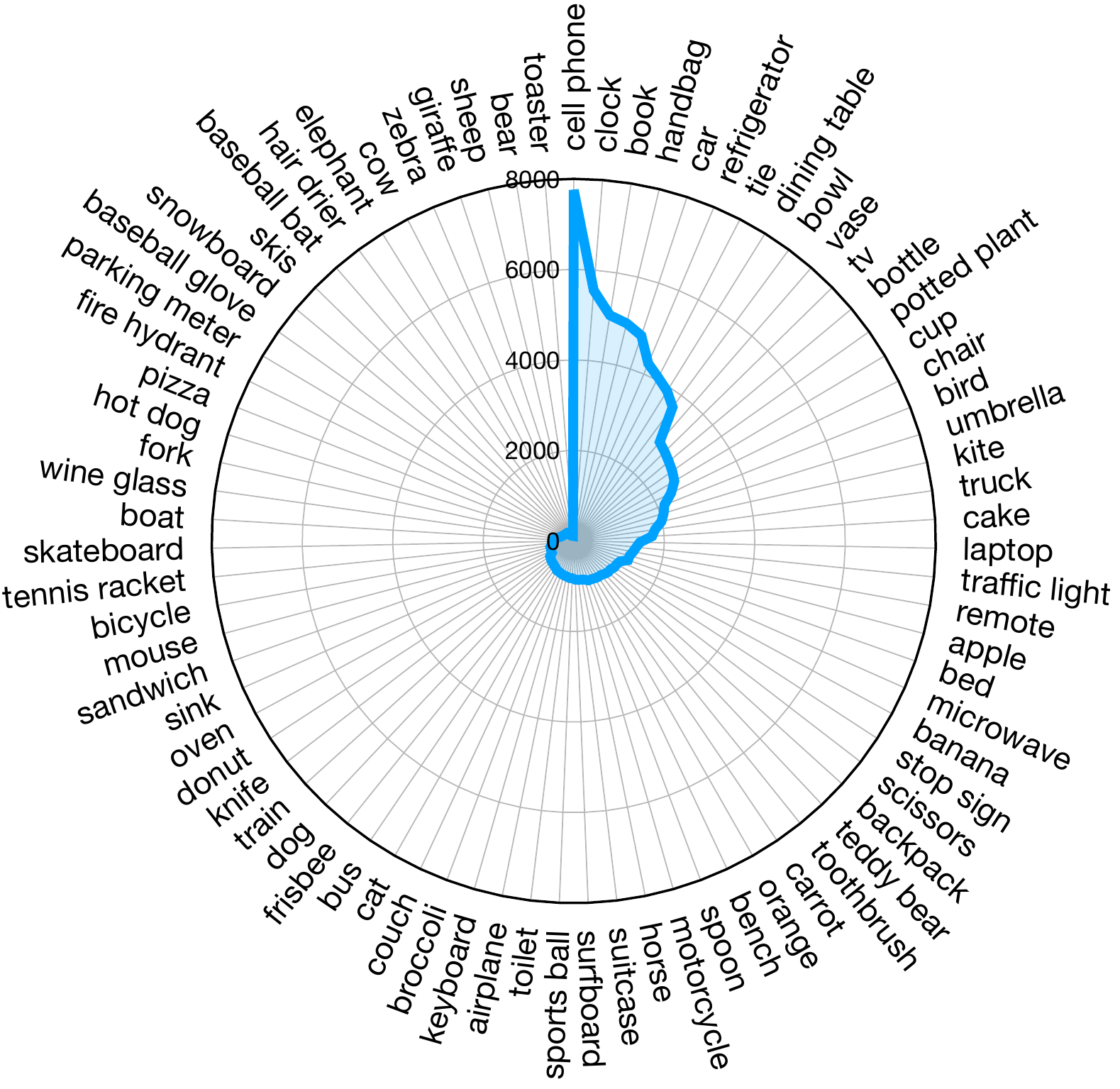}
		\caption{Distribution of the objects in our corpus. The objects are extracted via DETR and mapped to MS-COCO object categories. We remove \textit{perseon} object since it is too large for better illustration.}\label{label.image.class}
	\end{figure}

	\noindent
	\textbf{Question Classification:} To analyze the domain distribution of the labeled data, we use our in-house question-classification model to classify the questions in ChiQA. The result is shown on the left of Figure \ref{label.question.class}. We can see that the question in ChiQA spans a very wide range of domains, such as healthy, medical, education, etc. The domain distributions are even that none of them is dominant. The diversity of the domain makes the cross-modal learning on ChiQA more challenging since it should be more generalized.
	
	\noindent
	\textbf{Interrogative words:}  we also analyze the interrogative words of ChiQA. Different from other languages such as English, there are no apparent interrogative words such as \textit{who, when, what}, etc. Therefore, we randomly annotated some questions to extract their keywords such as  \begin{CJK}{UTF8}{gbsn}`如何'\end{CJK} (`how to'), \begin{CJK}{UTF8}{gbsn}`什么'\end{CJK} (`what'), etc. Then we map them to the interrogative words. For example, we map \begin{CJK}{UTF8}{gbsn}`的原因'\end{CJK} (`the reason of')  to the interrogative words \begin{CJK}{UTF8}{gbsn}`为什么'\end{CJK} (`why'). The question analysis of ChiQA is shown in Figure \ref{label.question.class}.

	\subsection{Image Analysis}
	We use images from the open-domain web that is very various, to analyze the characteristics of the images, we adopt an off-the-shelf object detection model to extract the objects in ChiQA. Concretely, we resort to the DETR \cite{carion2020end}, a Transformer-based end-to-end objective detection model that achieves state-of-the-art results on object detection, as our object extractor. The DETR is applied to the image where the corresponding questions are valid, the output is the distribution of common object categories defined by MS-COCO \cite{lin2014microsoft}, such as `\textit{person}', `\textit{dog}', etc. The distribution of the image objects is shown in Figure \ref{label.image.class}. We can see that more than 30 objects appear in more than 1,000 images.  The most common objects in ChiQA are `person', `cell phone', `car', etc. Which is in accordance with the question domain.
	
	\subsection{Reasoning Skills}\label{sec.reason}
	Previous analysis of VQA corpus mostly focused on vision-based understanding, such as object relation \cite{goyal2017making}, composition \cite{suhr2019corpus}, etc. However, we believe knowing the  \textbf{reasoning skills} requirement of VQA datasets is also important for the development of vision and language understanding systems. Inspired by the previous works of machine reading comprehension analysis \cite{sugawara2016analysis,sugawara2017evaluation}, we also identified 5 prerequisite reasoning skills for ChiQA\footnote{As indicated by \citet{sugawara2016analysis}, this categorization is somewhat empirical and provisional without very solid theoretical background.}.
	
	\textbf{1) Grounding:} the ability to locate and discriminate the target object mentioned in the question.
	
	\textbf{2) Reading:} the ability to recognize the text in the images with arbitrary fonts and layout.
	
	\textbf{3) Comparision:} is the knowledge to compare different objects with specific attributes, such as height, size, etc.
	
	\textbf{4)  Casual Inference:}  \cite{sugawara2017evaluation} is the knowledge about causality between the cause and effect.
	
	\textbf{5) Logical Knowledge:} understanding the predicate logic such as negation, conditionals, etc.
	
	Based on these prerequisite skills, we randomly sample 200 samples from the test set of ChiQA and annotate them, the statistics and examples are shown in Figure \ref{label.skill.class}. We can see that except for grounding, more than 80\% of the examples require \textit{reading}. This is in contrast with previous VQA datasets that mainly focus on non-textual objects. The reading skills and grounding skills are combined to facilitate a better understanding of the image in ChiQA.  Besides, we can see that a large number of examples require higher-level reasoning skills such as causal inference and logical reasoning, which are barely mentioned in previous VQA datasets. The prerequisite of high-level reasoning skills makes ChiQA challenging enough.
	
	\section{Experiments}
	\subsection{Evaluation Metrics}
	In ChiQA, the labels are three-level scores, so we adopt two types of metrics that are widely used:
	
	\textbf{NDCG@N}: When graded relevance judgments are available for a query, for instance in our ChiQA the three-point scale between zero to two. A popular incarnation of this metric is Normalized discounted cumulative gain. We report the NDCG scores at truncation levels 1, 3, and 5.
	
	\textbf{Mean Average Precision (MAP@N)} is another widely used metric in ranking results, it incorporates the trade-off between precision and recall. 
	
	\textbf{Accuracy/Precision/Recall/F1}: Besides the ranking based criteria, we also incorporate the classification based metric, which measures the \textit{absolute} score instead of the \textit{relative} score.  We consider two types of granularity: \textbf{Binary}, which takes 2 as positive and others (0,1) as negative. \textbf{Ternary}, a vanilla classification metric taking all three-level scores into account. Note that we use Macro-F1 in the ternary setting. For the binary result of precision and recall, we tune the threshold to obtain the best result.
	
	\subsection{Models}
	We evaluate a wide range of state-of-the-art vision and language understanding models on ChiQA. Inspired by previous works of image-text matching, we first represent the text and image by a text encoder and an image encoder, then their representations are merged to obtain a cross-modal representation. Finally, a prediction layer is applied on top of that representation to obtain the score. The encoders we consider are either pre-trained ($ \clubsuit $) or not ( $ \diamondsuit $):
	\begin{itemize}[wide, labelwidth=!, labelindent=0pt]
		\item \textbf{Random}$ ^\diamondsuit $: for each pair of test data, we simply choose a label from 2/1/0 randomly. The random baseline measures the lower bound of performance on ChiQA.
		\item \textbf{Majority}$ ^\diamondsuit $: every test sample is assigned with the score that appears mostly in training data. So every sample gets the same score and the rank-based metrics such as NDCG are meaningless.
		\item \textbf{LSTM+ResNet}$ ^\diamondsuit $: Before the prevailingness of large-scale pretraining, Long short term memory network \cite{Hochreiter1997LongSM} and residual networks \cite{he2016deep} are de-facto backbone for most language understanding and vision understanding systems.  We represent the question by a two-layer bi-directional  LSTM and the image by a 152-layer Resnet. Then their representations are fed to a cross-modal spatial attention layer proposed by \cite{zhu2016visual7w}, and finally, the mean pooled representations are used for prediction.
		
		\item \textbf{BERT+ViT}$ ^ \clubsuit  $: the text encoder and image encoder are initialized by Transformer-based BERT \cite{devlin2019bert} and ViT \cite{dosovitskiy2020image}, where both models are pre-trained on unlabelled data in a self-supervised way. In BERT+ViT, the presentation of text are initialized by a BERT$ _{\text{base}} $, and image are initialized by ViT-B/16 with weights pre-trained on ImageNet-21k. Then their representations are fed to the Transformer based self-attention layers. Finally, the representations of \texttt{<CLS>} token in BERT are used for prediction.
		
		\item \textbf{ALBEF }$ ^ \clubsuit  $:  is a large-scale vision and language representation learning has shown promising improvements on various vision-language tasks \cite{li2021align}. ALBEF also utilize the ViT and BERT to encode the text and images, however, their representations are \textit{aligned} by a contrastive learning objective before they are fused for cross-modal learning. In addition, the text encoder and multimodal encoder are both based on BERT model, making the architecture more efficient.

		\item \textbf{+Pretrain$ ^ \clubsuit  $}: In the raw model of BERT+ViT and ALBEF, only the text encoder and image encoder are pre-trained because the available cross-modal pre-training models are all based on English. For cross-modal pretraining on Chinese, we adopt Wukong \cite{gu2022wukong}, a large-scale Chinese pre-training dataset containing more than 100 million text-image pairs, and further pre-train BERT+ViT and ALBEF by a contrastive-learning method introduced in ALBEF \cite{li2021align}. 
		
		\item \textbf{Wenlan$ ^ \clubsuit  $} \cite{huo2021wenlan} is a large-scale multimodal pre-training model that was trained on more than 30 million text-images pairs and compact nearly one billion parameters. It outperforms both UNITER \cite{chen2020uniter} and OpenAI CLIP on various downstream tasks. Since the model are not publicly available so we adopt their public api\footnote{https://github.com/chuhaojin/WenLan-api-document} to obtain the representations of the question and the image, and use the cosine similarity as the output.
		
		\item \textbf{Human:} In addition to these models, we also ask two volunteers to obtain the human performance of ChiQA. Since the test set is very large, we only sample 100 of them to the volunteers to get an estimated result. 
		
	\end{itemize}
	
	\subsection{Common Setup and Objective}
	The hidden size of the LSTM is set to 512. For all image encoders, we resize the images with resolution 224 $ \times $ 224. The input question is truncated with the max length of 64.  The batch size was set to 256. We use the Adam \cite{Kingma2014AdamAM} optimizer with 400 warm-up steps and linearly decay the learning rate from $ 2\times10^{-5} $. $ \beta_1, \beta_2, \epsilon $ was set to 0.9, 0.999 and $ 10^{-8} $, respectively. Since the public available pretrained ALBEF models are based in English, we translate the questions in ChiQA into English to obtain the off-the-shelf result. We randomly select 2000 samples from the training set for development. We implement BERT$ _{\text{base}} $ and ViT-B/16 by Huggingface Transformers \cite{wolf2019huggingface}. All experiments are conducted through 8 A100 GPUs.
	
	\textbf{Objective:} For all model with fine-tuning, the output is a scalar value and we normlize the output with \textit{sigmoid} function. We also normalize the label 2, 1, 0 to 1, 0.5, 0, respectively, the final objective for fine-tuning is a pointwise binary cross-entropy loss:
	\begin{equation}\label{loss}
		\mathcal{L} = -\frac{1}{N} \sum_{i=1}^N y_i \cdot \log s_i +(1-y_i) \cdot \log (1-s_i) 
	\end{equation}
	where  $ y_i $ is the normalized labels and $ s_i $ is the normalized prediction. $ N $ is the size of data in a batch.
	
	\renewcommand\arraystretch{1.08} 
	\begin{table*}[]
		\setlength{\tabcolsep}{2.5pt}{
			\begin{tabular}{ll|ccc|ccc|cccc|cccc|}
				\cline{3-16}
				& \multicolumn{1}{c|}{} & \multicolumn{3}{c|}{NDCG} & \multicolumn{3}{c|}{MAP} & \multicolumn{4}{c|}{Ternary}         & \multicolumn{4}{c|}{Binary}          \\ \cline{3-16} 
				&                                        & 1      & 3      & 5                           & 1     & 3     & 5                          & Acc    & F1     & Precision & Recall                      & Acc    & F1     & Precision & Recall                      \\ \noalign{\hrule height 1.15pt}
				\multicolumn{1}{|l|}{\multirow{4}{*}{\rotatebox[origin=c]{90}{\small{w/o fine-tune}}}} & \multicolumn{1}{l|}{Random}            & 40.26  & 50.69  & \multicolumn{1}{c|}{59.81}  & 15.88 & 21.25 & \multicolumn{1}{c|}{22.14} & 34.30  & 32.06  & 34.25     & \multicolumn{1}{c|}{34.52}  & 62.53  & 21.40  & 15.34     & \multicolumn{1}{c|}{35.38}      \\
				\multicolumn{1}{|l|}{}                               & \multicolumn{1}{l|}{Majority}          & 41.26  & 49.25  & \multicolumn{1}{c|}{59.00}  & 17.02 & 21.58 & \multicolumn{1}{c|}{22.77} & 56.08  & 23.95  & 18.69     & \multicolumn{1}{c|}{33.33}  & 85.59  & N/A  & N/A     & \multicolumn{1}{c|}{N/A}      \\
				\multicolumn{1}{|l|}{}                               & \multicolumn{1}{l|}{ALBEF*}            & 43.95  & 53.62  & \multicolumn{1}{c|}{62.17}  & 16.17 & 21.54 & \multicolumn{1}{c|}{22.47} & 42.89  & 34.38  & 34.91     & \multicolumn{1}{c|}{34.47}  & 15.06  & 24.80  & 14.21     & \multicolumn{1}{c|}{97.72}      \\
				\multicolumn{1}{|l|}{}                               & \multicolumn{1}{l|}{Wenlan}            & 44.02  & 53.44  & \multicolumn{1}{c|}{62.36}  & 15.92 & 21.21 & \multicolumn{1}{c|}{22.43} & 37.94  & 33.21  & 39.89     & \multicolumn{1}{c|}{37.79}  & 49.31  & 26.76  & 16.90     & \multicolumn{1}{c|}{64.26}  \\ \hline
				\multicolumn{1}{|l|}{\multirow{5}{*}{\rotatebox[origin=c]{90}{\small{w/ fine-tune}}}}  & \multicolumn{1}{l|}{LSTM+ResNet}       & 42.10  & 51.60  & \multicolumn{1}{c|}{60.76}  & 15.88 & 21.34 & \multicolumn{1}{c|}{22.04} & 31.96  & 20.10  & 29.06     & \multicolumn{1}{c|}{33.36}  & 58.75  & 25.43  & 17.19     & \multicolumn{1}{c|}{48.81}   \\
				\multicolumn{1}{|l|}{}                               & \multicolumn{1}{l|}{BERT+ViT}          & 58.03  & 64.96  & \multicolumn{1}{c|}{69.83}  & 24.43 & 28.00 & \multicolumn{1}{c|}{28.17} & 60.50  & 53.37  & 54.40     & \multicolumn{1}{c|}{52.90}  & 81.80  & 46.19  & 40.24     & \multicolumn{1}{c|}{54.20}  \\
				\multicolumn{1}{|l|}{}                               & \multicolumn{1}{l|}{BERT+ViT+Pretrain} & 57.31  & 65.08  & \multicolumn{1}{c|}{70.03}  & 23.50 & 27.65 & \multicolumn{1}{c|}{27.97} & 61.23  & 53.15  & 56.96     & \multicolumn{1}{c|}{52.11}  & 82.21  & 44.50  & 40.42     & \multicolumn{1}{c|}{49.49}  \\
				\multicolumn{1}{|l|}{}                               & \multicolumn{1}{l|}{ALBEF}             & 58.96  & 65.97  & \multicolumn{1}{c|}{70.56}  & 24.60 & 28.50 & \multicolumn{1}{c|}{28.65} & 60.52  & 56.20  & \textbf{57.39}     & \multicolumn{1}{c|}{56.90}  & 84.37  & 46.72  & \textbf{45.91}     & \multicolumn{1}{c|}{47.56}  \\
				\multicolumn{1}{|l|}{}                               & \multicolumn{1}{l|}{ALBEF+Pretrain}    &\textbf{60.84} &\textbf{67.19} & \multicolumn{1}{c|}{\textbf{71.55}}  &\textbf{25.11}& \textbf{28.80}& \multicolumn{1}{c|}{\textbf{28.92}} &\textbf{63.25}  & \textbf{57.20}  & 57.06     & \multicolumn{1}{c|}{\textbf{57.48}}  & \textbf{82.24} & \textbf{47.77}  & 41.45     & \multicolumn{1}{c|}{\textbf{56.36}}  \\ \hline
				\multicolumn{2}{|c|}{Human}              & 73.33  & 77.55 & 80.54 & 30.00   & 32.33      & 32.15      & 81.85 & 80.54 & 86.36    & 77.59 & 95.17 & 77.88 & 100 & 63.77   \\ \noalign{\hrule height 1.15pt}
		\end{tabular}}
		\caption{Main result of different models on ChiQA. We report Macro-F1 in ternary classification.}\label{table.result}
	\end{table*}

	\subsection{Main Results}
	The result is shown in Table \ref{table.result}. We can see that directly applying the previous state-of-the-art cross-modal methods yield a poor performance, only achieving a negligible improvement over the random baselines. This implies that ChiQA is very hard and the models only involve the large-scale weakly supervised contrastive learning (ALBEF, Wenlan) may not able to discriminate the fine-grained information required for visual question answering. In addition, the poor performance of these models demonstrates that ChiQA is very different from the weakly supervised image-caption pairs. Since the image caption only captures the relevance, the ChiQA focuses on answerability.
	
	Furthermore, we observe that the additional cross-modal pretraining could bring some improvement but the margin is minor compared to the mono-modal pretraining. Therefore, the contrastive learning based cross-modal pretraining in ChiQA may not be as useful as they have achieved on other tasks, a better pretraining strategy should be explored.
	Finally, the model fine-tuned on ChiQA achieves a great improvement over the baselines, however, still far from human performance. So there is still a large room for improvement on ChiQA.
	
	\subsection{Ablation Study}
	%
	%

	\subsubsection{\textbf{Listwise v.s. Pointwise}} 
	All the models mentioned before are optimized by the pointwise objective, however, previous works on learning to rank demonstrate that the listwise ranking objective has superior performance \cite{ai2018learning}. Therefore, we replace the pointwise objective in equation \ref{loss} with ListNet, which is a listwise loss introduced in \cite{cao2007learning}, to enable the cross-image interaction, we concatenate all image features to a single sequence features and then fed them to the cross-modal encoder. The result is shown in Table \ref{table.obj}. We can see that the performance of listwise loss is not comparable with the pointwise loss, especially for the \textit{absolute} metrics such as accuracy and F1. Our findings suggest that ChiQA is more than just an image ranking problem, the system should predict the invariant answerability score to achieve good performance.
	\begin{table}[]
		\setlength{\tabcolsep}{2.7pt}{
			\begin{tabular}{|l|c|c|c|c|c|c|}
				\hline
				& NDCG@1 & MAP@1 & T-ACC & T-F1  & B-ACC & B-F1  \\ \hline
				Pointwise      & 58.03  & 24.43 & 60.50 & 53.37 & 81.80 & 46.19 \\ \hline
				Listwise       & 56.15  & 23.50 & 30.98 & 19.17 & 55.56 & 36.66 \\ \hline
		\end{tabular}}
		\caption{Experimental result of different training objective. T means ternary and B denotes binary.}\label{table.obj}
	\end{table}

	\subsubsection{\textbf{External Knowledge}} 
	Some previous works on cross-modal learning indicate that incorporating the external knowledge, such as objective detection \cite{li2020oscar}, could improve the downstream task a lot. We also consider two types of external knowledge for ChiQA:
	\begin{itemize}[wide, labelwidth=!, labelindent=0pt]
		\setlength{\itemsep}{0cm}
		\setlength{\parsep}{0.0cm}
		\setlength{\parskip}{0.0cm}  
		\renewcommand{\labelitemi}{\scriptsize$\blacksquare$}
		\item\textbf{object dection}:  most of the previous works on cross-modal learning resort to object detection to extract region-based image features and build the multimodal encoder thereon. We use DETR \cite{carion2020end}, a state-of-the-art object detector to predict boxes around both stuff and things classes on MS-COCO, then the region vectors extracted by DETR are treated as the input to the multimodal encoder.
		\item\textbf{optical character recognition (OCR)}: as we have shown in Sec. \ref{sec.reason}, more than 80\% of the questions in ChiQA require reading text in the image. Therefore, we use our in-house Chinese OCR system to extract the embedded text in the images, and append the extracted text to the question with a special \texttt{SEP} token, the concatenated texts are fed to the text encoder, then their mixed representations interact with the images to get the final score. 
	\end{itemize}
	
	The result of employing external knowledge is shown in Table \ref{table.external}, we can see that different types of the resources have different influences: When we only take the object representations to the model, the result even drops, therefore, the off-the-shelf object detection model may lose some information that is important for ChiQA. On the other hand, the enhancement of text information introduced by OCR could boost the performance a lot, so the \textit{reading} ability plays a very important role in VQA, the attention to the textual content, in combination with other content in the images should be further explored.

	\begin{table}[]
		\small
		\setlength{\tabcolsep}{2.8pt}{
			\begin{tabular}{|l|c|c|c|c|c|c|}
				\hline
				& NDCG@1 & MAP@1 & T-ACC & T-F1  & B-ACC & B-F1  \\ \hline
				BERT+ViT      & 58.03  & 24.43 & 60.50 & 53.37 & 81.80 & 46.19 \\ \hline
				BERT+DETR      & 55.56  & 22.95 & 57.38 & 52.00 & 78.32 & 41.98 \\ \hline
				BERT+ViT+OCR       &\textbf{64.90} & \textbf{27.73} & \textbf{65.98} & \textbf{60.05} & \textbf{84.97} & \textbf{53.14} \\ \hline
		\end{tabular}}
		\caption{Experimental results of employing different external knowledges. DETR is the state-of-the-art objective detection models, and OCR is the optical character recognition module to extract the text in the image.}\label{table.external}
	\end{table}
	
	\section{Conclustion}
	This paper introduces the ChiQA, a large-scale image-independent visual question answering dataset. In ChiQA, given the real-world user questions and related images, the task is to predict whether the image could answer the question. We adopt a very rigorous data inspection process to guarantee quality. In addition, we adopt an active learning based data selection process to improve the diversity and balance of the data.  The analysis demonstrates ChiQA is very challenging and requires a lot of high-level reasoning skills such as reading and comparison. Experimental results of many state-of-the-art language and vision understanding models demonstrate there is still a large gap between model and human performance, suggesting that there is still room for improvement in ChiQA.

	\bibliographystyle{ACM-Reference-Format}
	\bibliography{sample-base}

	\appendix
\end{document}